\documentclass[a4paper, 10pt, conference]{ieeeconf}      
\IEEEoverridecommandlockouts                              

\usepackage{lipsum}
\usepackage{comment}
\usepackage{graphicx}
\usepackage{hyperref}
\usepackage{todonotes}
\usepackage{comment}
\usepackage{cuted}
\usepackage{caption}
\usepackage[nolist]{acronym}

\acrodef{fair}[FAIR]{findable, accessible, interoperable and reusable}
\acrodef{vrb}[VRB]{Virtual Research Building}
\acrodef{neem}[NEEM]{Narrative Enabled Episodic Memories}
\acrodefplural{neem}[NEEMs]{Narrative Enabled Episodic Memories}

\title{\LARGE \bf
Open, Reproducible and Trustworthy Robot-Based Experiments with Virtual Labs and Digital-Twin-Based Execution Tracing
}

\author{Benjamin Alt$^{1,\dagger}$, Mareike Picklum$^{1}$, Sorin Arion$^{1}$, Franklin Kenghagho Kenfack$^{1}$ and Michael Beetz$^{1}$
\thanks{This work was supported by the German Research Foundation (DFG) as part of Collaborative Research Center 1320 ``EASE - Everyday Activity Science and Engineering'' and by the European Union’s Horizon 2020 research and innovation
programme under grant 101017089 (``TraceBot'').}
\thanks{$^{1}$AICOR Institute for Artificial Intelligence, University of Bremen}
\thanks{$^{\dagger}$Corresponding author: {\tt\small benjamin.alt@uni-bremen.de}}%
}

\begin{document}

\maketitle
\thispagestyle{empty}
\pagestyle{empty}

\begin{abstract}
We envision a future in which autonomous robots conduct scientific experiments in ways that are not only precise and repeatable, but also open, trustworthy, and transparent. To realize this vision, we present two key contributions: a semantic execution tracing framework that logs sensor data together with semantically annotated robot belief states, ensuring that automated experimentation is transparent and replicable; and the AICOR \ac{vrb}, a cloud-based platform for sharing, replicating, and validating robot task executions at scale. Together, these tools enable reproducible, robot-driven science by integrating deterministic execution, semantic memory, and open knowledge representation, laying the foundation for autonomous systems to participate in scientific discovery.
\end{abstract}

\section{INTRODUCTION}

The reproducibility crisis has emerged as a pressing challenge facing contemporary scientific research across disciplines \cite{bausellProblemScienceReproducibility2021}. Studies demonstrate that a substantial fraction of published results in the social, medical, natural, and engineering sciences cannot be replicated, undermining scientific inquiry and eroding trust in science \cite{andreolettiReplicabilityCrisisScientific2020}. Open science, the practice of sharing experimental protocols, code, data, tools, results and publications without barriers, has been identified as promising avenue toward addressing the reproducibility crisis and ensuring equitable, trustworthy scientific progress \cite{nationalacademiesofsciencesengineeringandmedicineOpenScienceDesign2018}.

We propose that integrating robots into scientific discovery processes not only accelerates research but fundamentally enhances the reproducibility and scientific rigor of experimental results and, if combined with \ac{fair} data principles and open computational infrastructure, can substantially boost the transformation toward open science. This transformation occurs through three key mechanisms. First, robots executing predefined protocols eliminate experimenter bias through mechanical repeatability and consistency, facilitating constant procedural conditions across trials. Second, robot code makes the operationalization of protocols transparent by providing computational implementations that can be preregistered and shared as supplementary materials, documenting exactly how protocols translate into concrete physical actions. Third, robots can generate comprehensive execution traces that provide ground-truth evidence of procedural rigor and the validity of results.

We present two key contributions toward enabling reproducible, robot-assisted scientific inquiry. The first is a semantic execution tracing framework which is a novel integration of low-level sensor logging, semantic scene annotations, and narrative reasoning traces in a single, unified data model. Unlike existing logging frameworks, ours captures \textit{why} actions were taken, \textit{how} perception decisions were made, and \textit{what} the robot believed at each point during task execution.


Our second contribution is the \textit{AICOR \acf{vrb}}\footnote{\url{https://vrb.ease-crc.org/}}, the first cloud platform that links containerized, deterministic robot simulations with semantically annotated execution traces. The \ac{vrb} provides open access to code, simulation environments, and data, enabling researchers worldwide to inspect, reproduce, and build upon each other's work.

These contributions form a foundation for the future development of autonomous robotic systems that can participate in scientific workflows. This work advances the long-term vision of open, traceable, and robot-supported science.

\section{RELATED WORK}

\subsection{Robots for Scientific Discovery}
Sparkes et al. define a ``Robot Scientist'' as a closed-loop system which ``generates hypotheses from a computer model of the domain, designs experiments to test these hypotheses, runs the physical experiments using robotic systems, analyses and interprets the resulting data, and repeats the cycle'' \cite{sparkes2010towards}. Such automatic, robot-enabled systems for accelerated scientific discovery have been proposed for a variety of domains such as genomics research \cite{sparkes2010towards}, pharmaceutical drug discovery \cite{williams2015cheaper}, multicomponent chemical formulation \cite{grizou2020curious} and materials science \cite{li2020toward}. While increasingly capable of end-to-end hypothesis generation, experimentation and evaluation, state-of-the-art robot scientist systems remain domain- and use case-specific. Truly autonomous, generalist scientists require both highly generalizable cognitive abilities and multi-purpose, highly flexible embodiments to perform scientific experiments in a variety of real-world contexts \cite{zhang2025scaling}. In doing so, robot scientists may, in fact, overcome several inherent limitations of human scientists \cite{kitano2016artificial}. One potential advantage is that robot scientists can make their cognitive processes explicit during experimentation, producing documentation that not only describes the states of the world during the experiment, but also the internal belief states of the experimenter. Coruhlu et al. ~\cite{9623348} developed a plan execution monitoring framework under partial observability that combines prediction, diagnosis, and explanation to enable autonomous self-checking. 

The CRAM cognitive architecture \cite{beetz2025robot,beetz2010cram} uses a semantic world model as the core representation for robot belief states, forming the basis for planning, reasoning and testing hypotheses about robot actions and world states. \acp{neem}~\cite{beetz2018know} persist robot belief states together with sensory percepts across robot actions. TraceBot\footnote{\label{tracebot}\url{https://www.tracebot.eu/}} tightly coupled a semantic digital twin with the robot's perception and plan executives, enabling the generation of comprehensive audit trails for pharmaceutical R\&D workflows \cite{mania2025imagisticreasoning}.

\subsection{Open Robotic Experimentation and Data Infrastructure}
Reproducibility in science hinges not only on rerunning code but on the faithful reconstruction of experimental conditions, data structures, and execution contexts. To address these challenges, containerization technologies have emerged as a foundation for computational reproducibility \cite{moreau2023containers}. Tools such as \textit{repo2docker}~\cite{repo2docker} and \textit{BinderHub}~\cite{binderhub2018} allow researchers to declaratively define software environments via Git repositories. These platforms automatically generate Docker images that encapsulate all dependencies, configurations, and scripts, enabling reproducible execution across diverse hardware setups. 

Beyond executable environments, reproducibility also requires capturing and sharing scientific data with sufficient provenance and structure. The \ac{fair} principles advocate for metadata-rich, machine-actionable data that supports long-term reuse \cite{scheffler2022fair}. Platforms like \textit{LiveDocs}~\cite{klein2024livedocscraftinginteractivedevelopment} extend this vision by allowing users to inspect and re-run code that generates scientific figures and results, and lower the barrier to reproducing and modifying scientific analyses. Similarly, \textit{MyBinder}~\cite{10365663} supports browser-based execution of Jupyter notebooks directly from Git repositories, facilitating widespread dissemination of executable research. Cloud robotics frameworks like \textit{Rapyuta}~\cite{6853392} provide secure, scalable environments for offloading robot computation to the cloud, enabling teams of robots to coordinate via shared knowledge repositories. However, these systems typically lack support for semantically structured data or fine-grained tracking of provenance. We propose to combine code access, simulation, and data access into one digital platform for reproducible robot-enabled experimentation.

\section{A SEMANTIC EXECUTION TRACING FRAMEWORK FOR ROBOT TASKS}


To address the fundamental challenge of generating comprehensive and interpretable execution traces for robot task executions, we present a semantic execution tracing framework that integrates three complementary technologies developed in the TraceBot project. This framework automatically captures not only low-level sensor data and robot commands, but also the high-level reasoning processes, perceptual interpretations and verification steps that occur during procedural execution.

\begin{figure*}
    \centering
    \includegraphics[width=\textwidth]{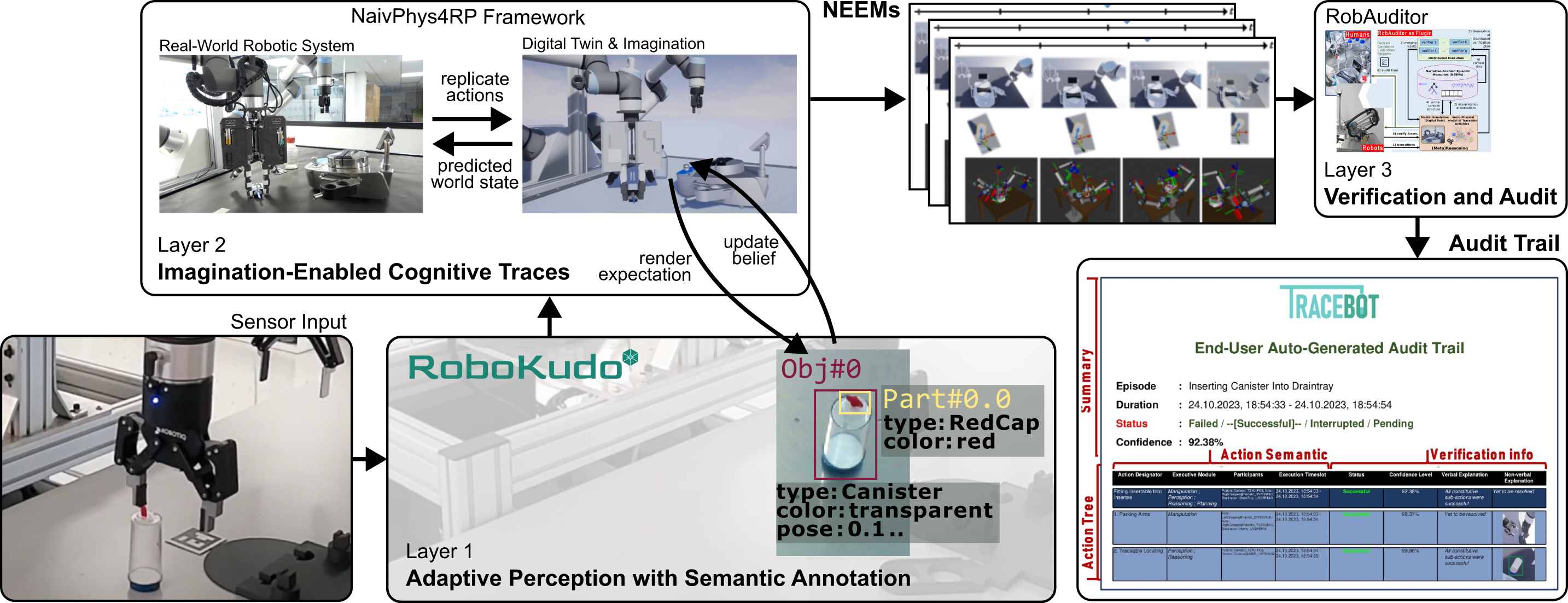}
    \caption{The TraceBot framework leverages a semantics-aware perception engine \cite{mania2024robokudo}, a semantic digital twin simulation and prospection framework \cite{10610037} and an imagination-enabled semantic verification system \cite{mania2025imagisticreasoning} to generate audit trails of robot actions, here for sterility testing.}
    \label{fig:tracing}
\end{figure*}

The semantic execution tracing framework operates through three interconnected layers that collectively provide multi-modal documentation of robot task executions (see \autoref{fig:tracing}).

\subsection{Layer 1: Adaptive Perception with Semantic Annotation}

The foundational layer employs the RoboKudo~\cite{mania2024open} perception framework, which models perception processes as Perception Pipeline Trees (PPTs) based on behavior tree semantics. Unlike monolithic perception systems, PPTs dynamically combine computer vision methods according to procedural requirements while maintaining complete traceability of perceptual decisions.
Each PPT consists of annotator nodes that represent individual vision methods (object detection, pose estimation, feature extraction) and control-flow nodes that orchestrate their execution. Annotators exchange information through a Common Analysis Structure (CAS) that accumulates semantic annotations throughout the perception process. This architecture enables the system to generate detailed traces documenting which perception methods were invoked, in what sequence, and with what intermediate results.
The framework captures multiple types of semantic information during perception:

1) Object hypotheses with confidence scores and classification rationales;
2) spatial relationships between detected entities with uncertainty estimates;
3) temporal sequences of perception events and their causal dependencies;
and 4) method selection justifications explaining why specific algorithms were chosen for given contexts.

For example, when detecting objects, the system documents not only the final detection results but also the cascade of perception methods attempted, the reasons for method selection, and the (low) confidence estimates that led to the selection of alternative perception methods. We hypothesize that interpretable logs of both perception results and metacognitive decisions about perception algorithms may increase trust in the cognitive system, particularly for hard and ambiguous perception problems such as the detection and classification of transparent objects.

\subsection{Layer 2: Imagination-Enabled Cognitive Traces}

The second layer integrates imagination-enabled perception capabilities that allow robots to generate and test hypotheses about the outcomes of their own task executions. This process is supported by high-fidelity simulations of semantic digital twins that mirror real-world laboratory environments, enriched with detailed object models and physics-based dynamics \cite{10610037}.

The process cycles through several stages:

\textbf{Hypothesis Generation:} Before executing actions, the system simulates anticipated outcomes in the digital twin environment.

\textbf{Action Synchronization:} Robot movements and manipulations are replicated in real-time within the semantic digital twin.

\textbf{Outcome Comparison:} Post-action observations are compared against simulated predictions using both pixel-level and semantic similarity metrics.

\textbf{Discrepancy Analysis:} When mismatches occur, the system generates detailed explanations of the differences and their potential causes.

This layer produces \textit{cognitive traces} that document the robot's reasoning about anticipated outcomes of the task executions~\cite{mania2025imagisticreasoning}. Cognitive traces are records of a robot's internal reasoning processes, such as hypotheses, predictions, decision-making steps, and causal explanations generated during task execution. They include visual comparisons between expected and actual results, explanations for task success or failure, and detailed analyses of object interactions during manipulation tasks. This process of \textit{cognitive emulation} is realized through the NaivPhys4RP framework~\cite{10610037}, which enables robots to reason about task execution contexts using commonsense knowledge about causality, physics, and object relationships. Commonsense knowledge in this context refers to structured, task-relevant knowledge about causal and physical properties of objects and actions (as axiomatized in the SOMA ontology \cite{bessler2021foundations} and derived, application-specific ontologies). Rules, such as ``containers must be open before pouring'' guide causal reasoning and interpretation of outcomes. The cognitive emulation process involves:

\textbf{Context Understanding:} Task execution protocols are converted into socio-physical knowledge graphs that capture object relationships, spatial configurations, and action sequences.

\textbf{Narrative Processing:} Task descriptions are interpreted through an Abstract Context Description Language (ACDL) that grounds natural-language instructions in domain ontologies.

\textbf{Causal Reasoning:} Explanations for observed phenomena are generated based on physics simulation and commonsense knowledge about object behaviors.

\textbf{Ontological Grounding:} Observations during task execution are linked to structured knowledge representations that enable semantic queries and automated analysis.

The resulting execution traces include narrative descriptions of task execution steps, causal explanations for observed outcomes, and explicit documentation of the knowledge and assumptions underlying robot decision-making.

\subsection{Layer 3: Context-Adaptive Verification, Recovery and Audit}

\begin{figure}
\centering
\includegraphics[width=0.49\textwidth]{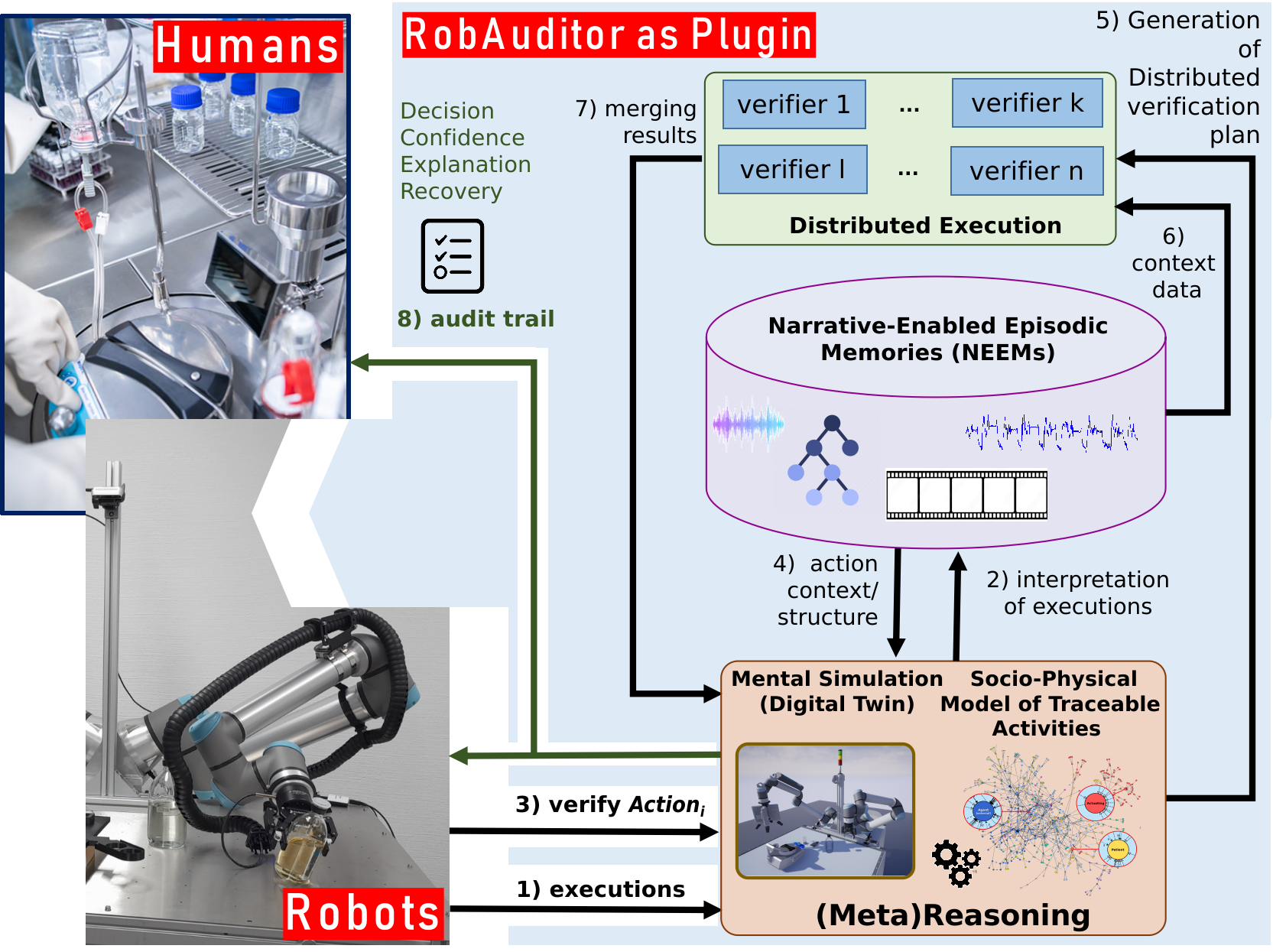}
\caption{RobAuditor: Context-adaptive task verification, recovery and audit \cite{kenghagho2025a}.}
\label{fig:robauditor}
\end{figure}

The flexibility afforded by robots enables the automation of increasingly complex procedures in increasingly unstructured environments, but also introduces novel, hard-to-detect ways in which task executions can fail. To ensure the integrity of the procedures, we introduced RobAuditor~\cite{kenghagho2025a} (see Figure \ref{fig:robauditor}), a plugin-like framework for context-aware and -adaptive task verification planning and execution, failure recovery, and audit trail generation.

\textbf{Workflow.} The RobAuditor workflow is illustrated in Figure \ref{fig:robauditor}. For plugging RobAuditor into the robot, a formal query-based language is provided for their interactions. (1) The tracer presented in Layer 1 and Layer 2 makes use of the interface to send execution traces to RobAuditor, (2) then RobAuditor interprets these traces into a comprehensive story, grounded in an established ontology (SOMA \cite{bessler2021foundations}) and persistently stored as \acp{neem} of the robot activities. (3) Any time (online/offline), a task verification query is issued, (4) RobAuditor's metareasoner will access the context (\acp{neem}+SOMA+DT) of the task, (5) then generate based on the context a distributed verification pipeline, made out of reasoning units called verifiers (competence domains and implementations defined in SOMA), (6) which will then be executed in a distributed manner. (7) As the pipeline executes, reasoning units can make use of RobAuditor's interface to access the context (e.g., what is the diameter of the object?). (8) RobAuditor's metareasoner synthesizes the final verification result from all reasoning units with (potentially) a recovery plan in case of failure. Note that each reasoning unit as well as the metareasoner returns for this verification a quadruplet $(D_s,C_f,E_d,E_r)$, denoting a boolean verification decision, a decision confidence, a decision explanation, and possibly a recovery plan (SOMA abstractly categorizes failures and defines corresponding recovery strategies), respectively. Eventually, the audit trail is generated as a concise documentation of the robot activities. We refer to \cite{kenghagho2025a} for a detailed overview and evaluation of the TraceBot framework on a sterility testing usecase \cite{remazeilles2022robotizing}.

\subsection{Discussion}
This semantics- and simulation-driven approach to execution tracing provides several methodological advantages for reproducible robot science. The framework generates complete documentation of robot reasoning processes, eliminating the black-box problem that many automated systems suffer from. Beyond documentation, semantic and cognitive traces allow researchers to understand not just \textit{what} the robot did, but \textit{why} it made specific decisions and how it arrived at execution outcomes. In future work, we are investigating imagination-enabled traces for automated formal verification of task execution, enabling detection of protocol deviations or unexpected outcomes in real time.

\begin{figure}
    \centering
    \includegraphics[width=1\linewidth]{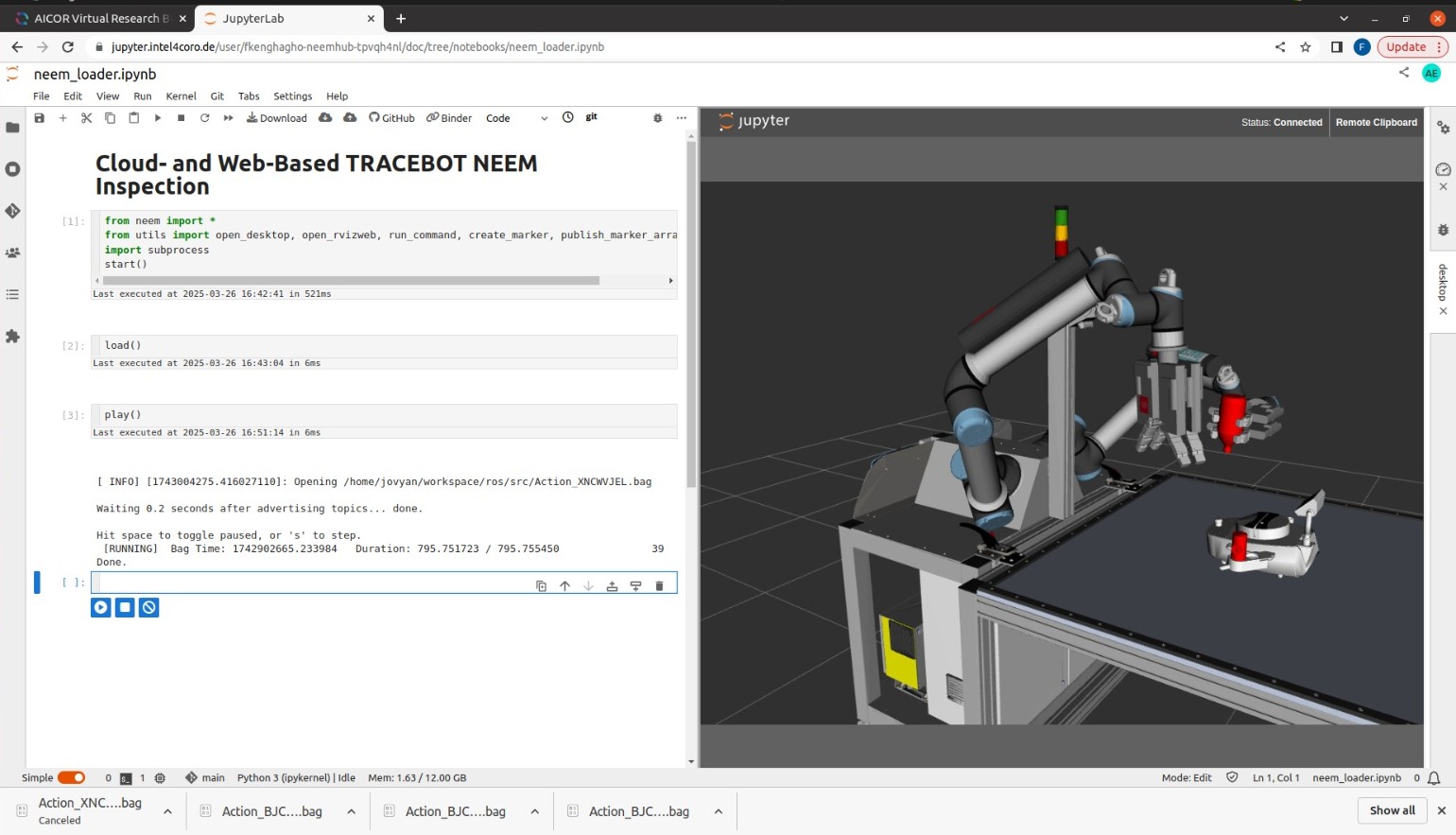}
    \caption[Replaying \acp{neem} in the VRB.]{Replaying \acp{neem} in the VRB.\footnotemark}
    \label{fig:vrb-neem}
\end{figure}

The modular architecture allows the framework to be extended with new perception methods, reasoning capabilities, or domain-specific knowledge without requiring complete system redesign. Consequently, it can accommodate task executions of varying complexity, from simple manipulation tasks to multi-step protocols involving complex object interactions. Through this integrated approach, the semantic execution tracing framework enables a new level of task-level documentation that supports both immediate reproducibility and long-term scientific analysis of robot-executed procedures. All system components are available open-source\footnote{\url{https://gitlab.com/tracebot}}\footnote{\url{https://robokudo.ai.uni-bremen.de}}\footnote{\url{https://github.com/NaivPhys4RP}}, and exemplary execution traces can be replayed and examined in the VRB (see \autoref{fig:vrb-neem}).

\section{VIRTUAL ROBOT LABORATORIES FOR OPEN, REPRODUCIBLE SCIENCE}

The \ac{vrb}\footnote{\url{https://vrb.ease-crc.org/}} provides a cloud-based infrastructure for hosting virtual laboratories that encapsulate complete procedural setups, enabling bit-level reproducibility and facilitating collaborative, robot-based experiments across the global research community.
\footnotetext{\url{https://binder.intel4coro.de/v2/gh/fkenghagho/NeemHub/HEAD?labpath=notebooks/neem_loader.ipynb}}

\subsection{Containerized Architecture for Computational Reproducibility}

The \ac{vrb} implements reproducibility through a multi-layered containerization architecture utilizing Docker containers. Each virtual laboratory instantiates as an isolated container embedding the complete CRAM 2.0 cognitive robotics software stack \cite{beetz2025robot}, including ROS, the Multiverse multi-backend simulation environment \cite{Multiverse}, the PyCRAM robot programming language \cite{dech2024pycram}, the KnowRob robot knowledge representation and reasoning engine \cite{beetz2018know}, as well as any optional domain-specific software. Container images are constructed from version-controlled Dockerfiles stored in Git repositories, ensuring precise specification of software dependencies, library versions, and system configurations.

The platform leverages BinderHub \cite{binderhub2018} for automated container image construction and deployment. When researchers commit code to public Git repositories, BinderHub automatically rebuilds Docker images incorporating all dependencies specified in the repository's Dockerfile. This process generates immutable container images with cryptographic hashes, providing verifiable computational environments for execution. The resulting containers execute identically across heterogeneous hardware platforms, eliminating variability introduced by different operating systems, library versions, or hardware configurations.

Importantly, the \ac{vrb}’s Docker container images are not solely accessible via the platform’s web interface but can also be downloaded for local deployment. This enables researchers to instantiate fully interactive, high-fidelity VR simulations and high-performance cognitive robotics simulation on their own computational infrastructure. Local execution facilitates resource-intensive workloads and immersive user interaction while preserving the fidelity and reproducibility of the experiment setup, thus affording maximum flexibility for offline analysis, customized parameter tuning, and integration within heterogeneous research environments.

\begin{figure}
    \centering
    \includegraphics[width=1\linewidth]{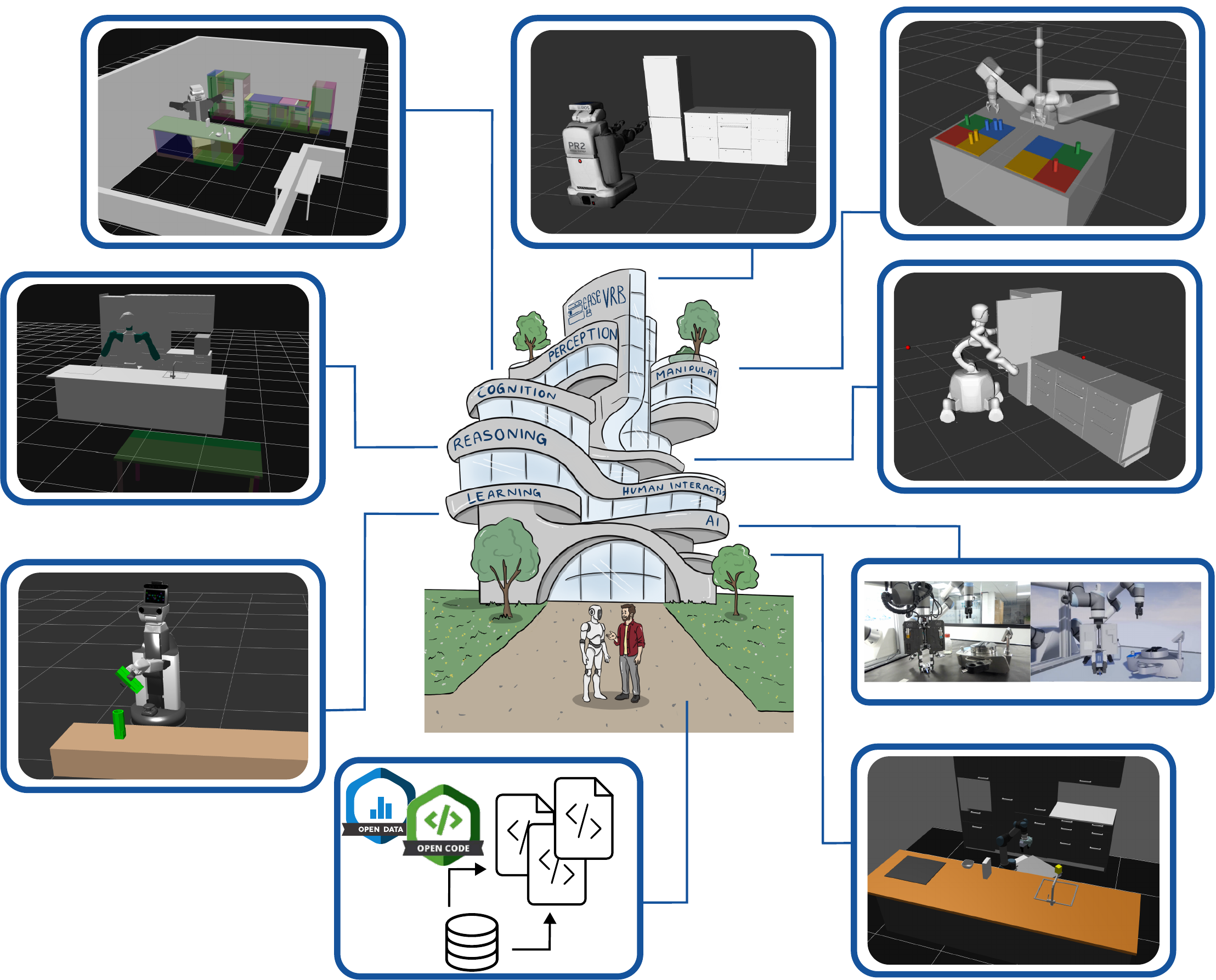}
    \caption{The \ac{vrb} offers a wide range of existing laboratories to explore, enabling users worldwide to access open-source code and data from open cloud repositories. Its collaborative approach supports key aspects of open science, such as the reproducibility of robot task executions.}
    \label{fig:openscience}
\end{figure}


Local \ac{vrb} deployments can take advantage of the Multiverse~\cite{Multiverse} simulation framework, which provides a unified interface to multiple simulation engines including MuJoCo \cite{todorov2012mujoco}, Bullet Physics \cite{coumans2015bullet}, NVIDIA Isaac Sim, and Gazebo \cite{koenig2004design}. Each simulation backend exhibits different deterministic properties: MuJoCo provides deterministic forward dynamics for continuous control tasks, Bullet Physics offers deterministic rigid body dynamics with configurable solver parameters, and Gazebo implements deterministic discrete-time simulation with controllable integration schemes. Researchers select simulation backends based on their specific performance and determinism requirements, with Multiverse ensuring consistent Universal Scene Description (USD) data structures across different simulators \cite{nguyen2024translating}.

\begin{figure}
    \centering
    \includegraphics[width=1\linewidth]{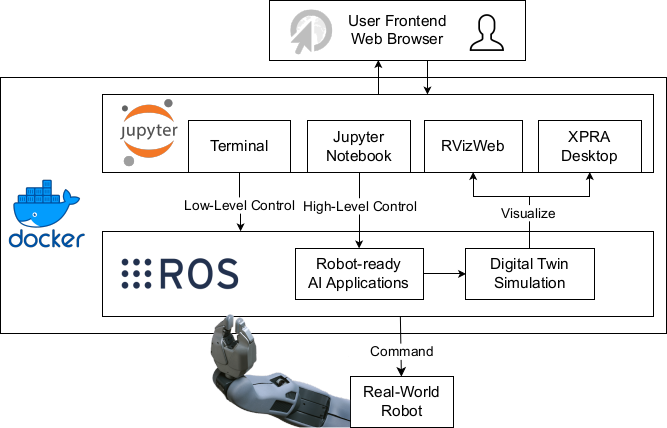}
    \caption{The VRB deploys virtual laboratories as sandboxed, individually deployable Docker containers. The pre-installed robotics software stack can be extended by arbitrary additional domain- or application specific software.}
    \label{fig:architecture-docker}
\end{figure}

\subsection{Data Provenance and Episodic Memory Architecture}

The \ac{vrb} integrates with \textit{NEEMHub}~\cite{beetz2020neem}, a distributed knowledge service implementing the \ac{neem} data model for comprehensive, semantically rich execution traces. NEEMs capture complete episodes as timestamped MongoDB documents containing multimodal sensor data, robot state trajectories, environmental object configurations, and semantic event annotations. The temporal structure of \acp{neem} enables precise reconstruction of procedural timelines with microsecond-level timestamp resolution.

The \ac{neem} data model implements a three-layer architecture: raw sensor data streams, symbolic state representations, and semantic annotations. Raw data includes RGB-D camera feeds, joint position encoders and force-torque sensor readings. Symbolic representations encode discrete state transitions including object poses, robot configurations, and task execution states in terms of the SOMA ontology \cite{bessler2021foundations}. Semantic annotations provide high-level descriptions of procedural events, task goals, and outcome assessments using standardized knowledge representation languages including OWL and SWRL.

The NEEMHub uses content-addressable storage (cryptographic hashes) of \ac{neem} documents to ensure reproducibility and trust in stored task executions, ensuring any change in data is detectable. This guarantees that execution traces are immutable and verifiable, which is essential  when procedures are replicated across different sites or over long time periods.
To improve usability, we extended the previous Prolog-based query interface with a Python-native solution using the PyCRAM framework \cite{dech2024pycram}. This new approach integrates a Pythonic Object-Relational Mapping (ORM) layer for querying the MongoDB backend, significantly lowering the entry barrier for users unfamiliar with logic programming. Additionally, the framework now automatically logs all robot actions, belief states and object interactions to enable tracking, debugging, and reproducibility.

The temporal fidelity of \ac{neem} replay is maintained through MongoDB's timestamp ordering and the platform's deterministic event scheduling mechanisms.

\subsection{Deterministic Execution and Validation Mechanisms}

The reproducibility of robot actions in the \ac{vrb} depends on the deterministic properties of the underlying software components. The Giskard motion planner~\cite{9982245} implements deterministic trajectory optimization using sequential quadratic programming with fixed random seeds, ensuring identical motion plans for identical initial conditions and constraints. The PyCRAM plan executive~\cite{dech2024pycram} provides deterministic task execution through symbolic planning algorithms with reproducible search strategies and tie-breaking rules.

Recorded \acp{neem} contain a serialized timeline of the robot’s belief state alongside sensor data and action histories, enabling precise replay of past task executions (see Figure \ref{fig:vrb-neem}). This allows researchers not only to replicate robot behavior exactly as it occurred, but also to inspect and modify specific aspects of the execution for comparative or diagnostic analysis. By preserving the full internal and external context of the robot's execution, the platform supports rigorous post hoc validation, traceability, and reasoning over autonomous decision-making processes.

The \ac{vrb} implements \textit{semantic validation} mechanisms through automated comparison of \ac{neem} episodes by comparing their structured, meaning-based (semantic) representations (e.g. ontological task descriptions) against expected or reference models.
High-level task outcomes are validated by comparing semantic annotations between original and reproduced task executions using graph isomorphism algorithms applied to task execution trees. This approach enables validation of task execution reproducibility even when low-level robot motions exhibit minor variations due to numerical precision limitations or simulator differences.

\subsection{Knowledge Representation and Domain-Specific Validation}

The VRB's knowledge representation framework enables domain-specific validation criteria through extensible ontology systems. The SOMA ontology provides core concepts for robotic manipulation tasks, while researchers can integrate domain-specific ontologies for specialized experimental validation \cite{bessler2021foundations}. For materials science procedures, researchers might implement ontologies describing crystal structures and phase transitions. For biological research, specialized ontologies could encode protein folding states and molecular interactions.
Semantic rule engines implemented in the Semantic Web Rule Language (SWRL) enable automated quality assessment of data. Rules can specify validity conditions such as ``successful grasping requires contact forces exceeding threshold values'' or ``navigation tasks must maintain minimum clearance distances from obstacles.'' These rules execute automatically during task execution, providing quantitative assessments of data quality and task success criteria.
Actionable Knowledge Graphs (AKGs) extend the utility of knowledge representation by linking object knowledge to both environment and action knowledge in a way that supports decision-making and automated behavior across various agents \cite{kumpel2024actionable}. For instance, a product knowledge graph supports omni-channel shopping assistance by integrating product data with contextual and spatial information, accessible by smartphones, smart glasses, or robots \cite{kumpel2023robotic}. Similarly, a food cutting knowledge graph allows robots to autonomously perform variations of cutting tasks by incorporating object affordances and web-acquired procedural knowledge \cite{dhanabalachandran2021cutting}. These AKGs empower systems to reason over task-relevant knowledge and execute context-appropriate actions, demonstrating the integration of semantic representation with real-world functionality.

The knowledge representation system supports logical reasoning over data using description logic inference engines. Researchers can formulate hypotheses as logical queries over \ac{neem} databases, enabling systematic testing of scientific hypotheses across large datasets. This capability supports meta-analyses and systematic reviews that would be computationally intractable with traditional approaches.

\subsection{Procedure Sandboxing and Parallel Execution}

The \ac{vrb} upholds principles of transparent collaboration and reproducibility. Lab maintainers are tasked with ensuring that all code is systematically committed to version control systems such as Git, which supports traceability and accountability. While operators can interactively modify and test code within the lab environment, these changes remain transient. Any permanent modifications to configurations or protocols must be formally committed and managed through version control.

The platform supports robustness and flexibility through strong procedure sandboxing. 
Each task execution runs in its own isolated container, ensuring that changes made by one user do not affect others’ ongoing work. This isolation facilitates safe experimentation and reproducibility, allowing users to modify or extend procedures without fear of unintended side effects.
The platform implements fault tolerance through Kubernetes' self-healing mechanisms, automatically restarting failed containers and rescheduling virtual laboratories on healthy compute nodes. Persistent volumes ensure data survival across container failures, while distributed storage systems provide data redundancy and high availability.

\begin{figure}
    \centering
    \includegraphics[width=1\linewidth]{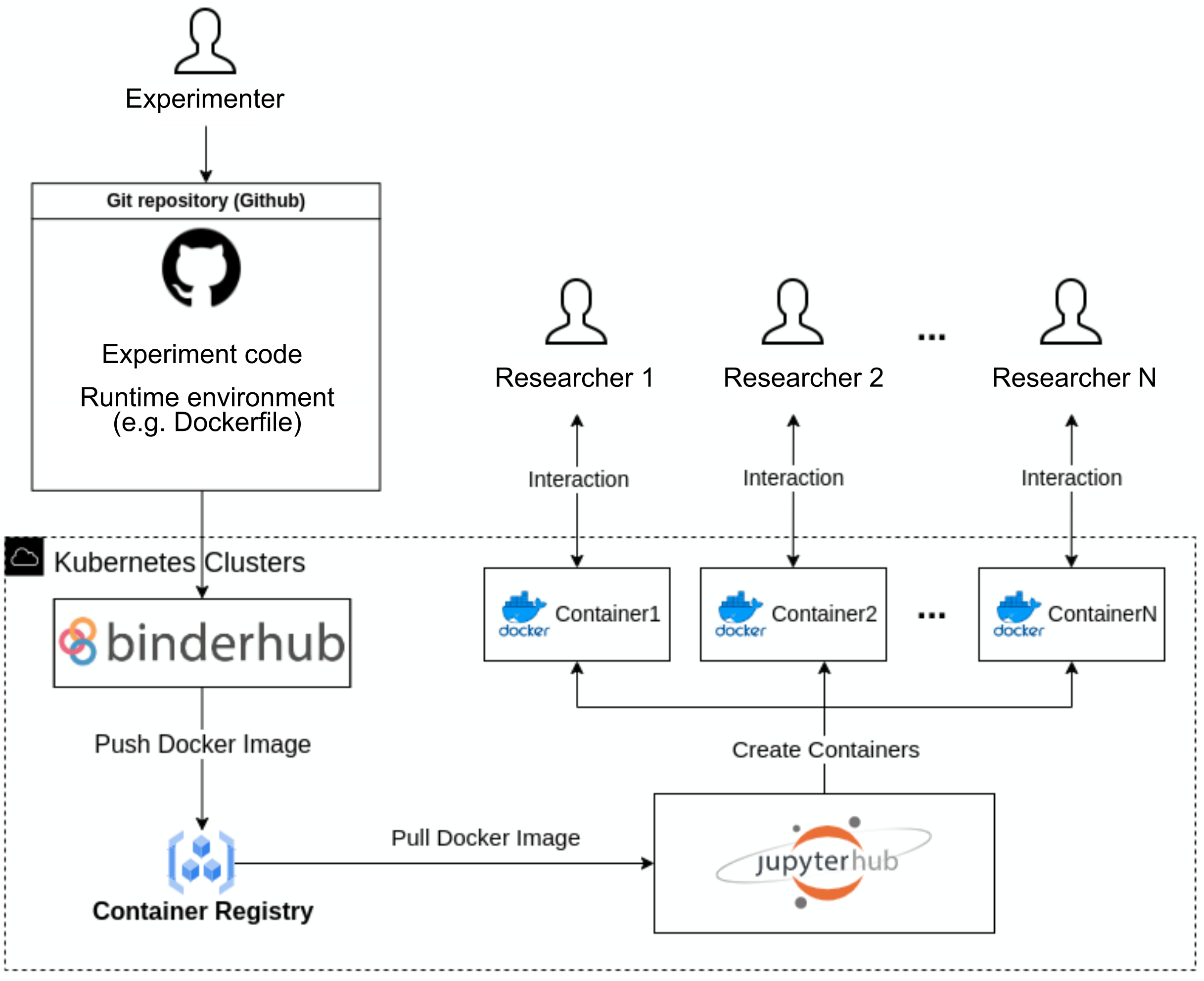}
    \caption{The VRB allows virtual laboratory maintainers to share version-controlled code along with the environment required to run it. Researchers can inspect, reproduce and interact with robot task executions in individual, sandboxed environments.}
    \label{fig:architecture-cloud}
\end{figure}

\subsection{Discussion}

\subsubsection{Limitations and Numerical Considerations}

The \ac{vrb}'s reproducibility guarantees are subject to several technical limitations. Floating-point arithmetic in physics simulators can exhibit platform-dependent behavior due to differences in CPU architectures, compiler optimizations, and mathematical library implementations. While IEEE 754 floating-point standards provide consistency within specific hardware platforms, cross-platform reproducibility may require additional validation.
Non-deterministic algorithms including certain machine learning methods, genetic algorithms, and simulated annealing procedures require careful treatment to achieve reproducibility. The platform provides facilities for deterministic pseudo-random number generation, but researchers must explicitly manage randomness sources within their code.
Real-time constraints in robotic systems can introduce timing-dependent behavior that affects task execution outcomes. The \ac{vrb} addresses this through deterministic simulation scheduling and configurable time step control, but researchers must validate that their task executions are robust to small timing variations.

\subsubsection{Implications for Computational Scientific Discovery}

The \ac{vrb}'s virtual laboratory infrastructure enables new paradigms for computational validation in robotics-assisted scientific discovery. The platform's ability to capture, version, and replay complete episodes creates opportunities for systematic reproducibility studies, large-scale parameter sweeps, and collaborative validation across research institutions. This infrastructure supports the development of more rigorous scientific methodologies for procedures involving complex robotic systems and autonomous agents.
The platform's integration of symbolic knowledge representation with low-level sensor data enables novel approaches to scientific hypothesis testing and discovery. Researchers can formulate scientific hypotheses as logical queries over large \ac{neem} databases, testing theoretical predictions against empirical observations collected from robotic task executions. This capability bridges the gap between theoretical modeling and empirical validation in scientific research involving embodied AI systems.

\section{CONCLUSION}

We presented a framework and infrastructure for enabling open, reproducible, and trustworthy robot-based research through semantic execution tracing and virtual laboratories. Our semantic execution tracing framework captures not only raw sensor and control data but also interpretable and structured representations of robot perception, beliefs, and reasoning processes. This multi-layered traceability goes beyond traditional logging and supports deep insight into robot behavior, facilitating transparent, repeatable robotic task procedures.

Building on this foundation, the \ac{vrb} provides a cloud-based platform for deploying, sharing, and replicating robotic task executions at scale. By combining containerized environments, deterministic simulation, structured data storage, semantically annotated task records, and ontological validation, the \ac{vrb} addresses critical barriers to reproducibility in robotics-driven research. It aligns with the FAIR principles and supports transparent evaluation, domain-specific validation, and collaborative reuse through its integration with knowledge representation and episodic memory structures.

We have implemented and demonstrated a reproducibility pipeline that combines deterministic robotic execution, semantic and cognitive trace logging and cloud-based sharing through the \ac{vrb}.
This allows researchers to reproduce not only the outcome of a task execution but analyze the internal decision making of the robot. We believe this addresses addresses a critical gap in reproducible robotics and provides a practical foundation for open, trustworthy, robot-enabled science.





\bibliographystyle{IEEEtran}
\bibliography{bibliography.bib}

\addtolength{\textheight}{-12cm}   

\end{document}